\documentclass[11pt]{article}

\usepackage[final]{acl}

\usepackage{times}
\usepackage{latexsym}

\usepackage{microtype}
\usepackage{hyperref}
\usepackage{url}
\usepackage{booktabs}
\usepackage{amsmath}
\usepackage{amssymb}
\usepackage{tabularx}
\usepackage{makecell}

\usepackage{colortbl}    

\usepackage{times}
\usepackage{float}
\usepackage{latexsym}
\usepackage{booktabs}
\usepackage{amsmath}
\usepackage{hyperref}    
\usepackage{amssymb}     
\usepackage{tipa}
\usepackage{multirow}
\usepackage{xcolor}
\definecolor{mypink}{rgb}{0.9686274,0.882352,0.86666}
\definecolor{mygreen}{rgb}{0.819607,0.890196,0.760784}

\usepackage{latexsym}
\usepackage{graphicx}
\usepackage{float}
\usepackage{colortbl}
\usepackage{multirow}
\usepackage{booktabs}
\usepackage{subcaption}
\usepackage{rotating}
\usepackage{placeins}
\usepackage{comment}
\usepackage{pifont}

\usepackage[utf8]{inputenc}
\DeclareUnicodeCharacter{1E24}{\d{H}}
\DeclareUnicodeCharacter{1E24}{\d{H}}
\usepackage{lineno}

\definecolor{darkblue}{rgb}{0, 0, 0.5}
\hypersetup{colorlinks=true, citecolor=darkblue, linkcolor=darkblue, urlcolor=darkblue}
\usepackage[T1]{fontenc}

\newcommand{\cmark}{\ding{51}} 

\definecolor{bestresult}{HTML}{D4EDDA}

\title{From Traditional Taggers to LLMs: A Comparative Study of POS Tagging for Medieval Romance Languages}

\author{
  Matthias Schöffel$^{1}$~~~
  Esteban Garces Arias$^{2,3}$\\
    $^{1}$Bavarian Academy of Sciences (BAdW), Munich, Germany\\
  $^{2}$Department of Statistics, LMU Munich, Germany\\
  $^{3}$Munich Center for Machine Learning (MCML), LMU Munich, Germany\\
  \texttt{matthias.schoeffel@badw.de} \\
  \texttt{esteban.garcesarias@stat.uni-muenchen.de}
}

\begin{document}
\maketitle
\begin{abstract}
Part-of-speech (POS) tagging for Medieval Romance languages remains challenging due to orthographic variation, morphological complexity, and limited annotated resources. This paper presents a systematic empirical evaluation of large language models (LLMs) for POS tagging across three medieval varieties: Medieval Occitan, Medieval Catalan, and Medieval French. We compare traditional rule-based and statistical taggers with modern open-source LLMs under zero-shot prompting, few-shot prompting, monolingual fine-tuning, and cross-lingual transfer learning settings.

Experiments on historically grounded datasets show that LLM-based approaches consistently outperform traditional taggers, with fine-tuning and multilingual training yielding the largest improvements. In particular, cross-lingual transfer learning substantially benefits under-resourced varieties, while targeted bilingual training can outperform broader multilingual configurations for specific target languages. The results highlight the importance of linguistic proximity and dataset characteristics when designing transfer strategies for historical NLP.

These findings provide empirical insights into the applicability of modern neural methods to medieval text processing and provide practical guidance for deploying LLM-based POS tagging pipelines in digital humanities research. All code, models, and processed datasets are released for reproducibility\footnote{\url{https://github.com/msch38/medieval-romance-pos/tree/main}}.
\end{abstract}

\section{Introduction}

The computational analysis of historical texts plays an increasingly important role in digital humanities, enabling large-scale investigation of linguistic change, cultural transmission, and textual variation. A fundamental prerequisite for such analyses is reliable linguistic annotation, with part-of-speech (POS) tagging serving as a key preprocessing step for downstream tasks including syntactic parsing, semantic analysis, and diachronic linguistic modeling \citep{piotrowski2012natural, ehrmann2020named}.

\begin{figure*}[h]
    \centering
    \begin{minipage}{0.45\textwidth}
        \centering
        \includegraphics[width=\textwidth]{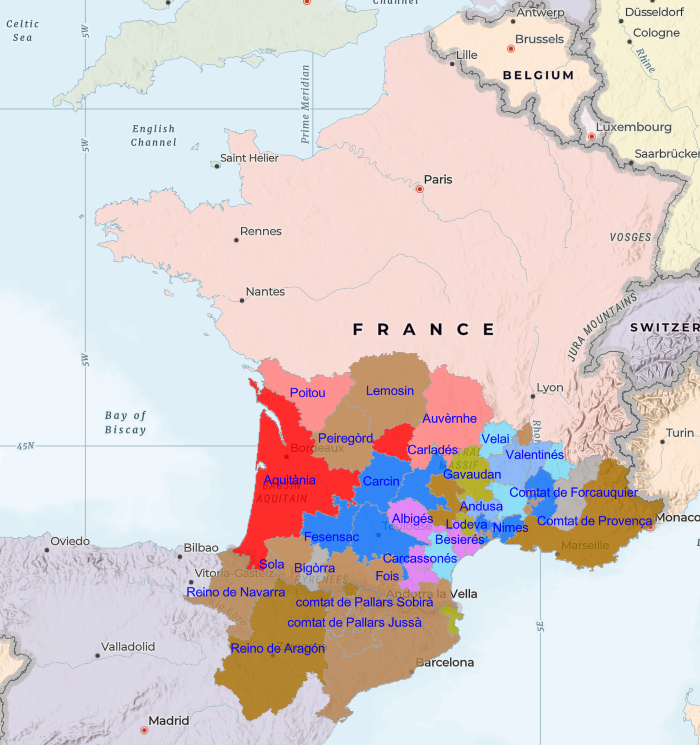}
\subcaption{Map of Medieval Occitan and Medieval Catalan variations, 13th century \citep{trobeu2025}.}
        \label{fig:map}
    \end{minipage}
    \hfill
    \begin{minipage}{0.48\textwidth}
        \centering
        \small
        \begin{tabular}{p{2.4cm}p{3.6cm}}
        \toprule
        \textbf{Spelling variants} & \textbf{Modern/Known spelling} \\
        \midrule
        \multicolumn{2}{l}{\textbf{Medieval French}} \\
        \textit{deffendre}   & défendre (engl. 'to defend') \\
        \textit{joinncture}  & jointure (engl. 'knuckle') \\
        \textit{sun}   & son (engl. 'his') \\
        \textit{pruz}   & preux (engl. 'brave') \\
        \midrule
        \multicolumn{2}{l}{\textbf{Medieval Catalan}} \\
        \textit{ssaber}   & saber (engl. 'to know') \\
        \textit{Ffran\c{c}a}  & Fran\c{c}a (engl. 'France') \\
        \textit{hòmens}   & homes (engl. 'men') \\
        \textit{jóvens}   & joves (engl. 'youth') \\
        \midrule
        \multicolumn{2}{l}{\textbf{Medieval Occitan}} \\
        \textit{deceplina}   & disciplina, disiplina, desiplina (engl. 'discipline') \\
        \textit{falssa}  & fals (engl. 'false') \\
        \textit{liech}/\textit{lech}   & lloc (engl. 'place') \\
        \textit{fuoc}/\textit{foc}   & foc (engl. 'fire') \\
        \bottomrule
        \end{tabular}
        \subcaption{Spelling characteristics across medieval language variations.}
        \label{fig:table}
    \end{minipage}
\caption{Geographic distribution and spelling characteristics of medieval Romance languages (13th century). Left: regional variation of Medieval Occitan and Medieval Catalan. Right: spelling variants across Medieval French, Medieval Catalan, and Medieval Occitan.}
    \label{fig:combined}
\end{figure*}

POS tagging for Medieval Romance languages presents particular challenges. These varieties exhibit substantial orthographic instability, dialectal diversity, and complex morphological systems, while annotated corpora remain comparatively small and domain-specific \citep{schoeffel2025modernmodelsmedievaltexts}. An overview of the geographic distribution as well as the spelling characteristics in all languages is shown in Figure~\ref{fig:combined}. As a result, computational tools originally developed for modern standardized languages often show limited robustness when applied to medieval textual material.

Traditional approaches to historical POS tagging have relied on rule-based systems and statistical models adapted from modern resources. While tools such as COLaF\footnote{\url{https://colaf.huma-num.fr/deucalion/occ-cont}} and UDPipe\footnote{\url{https://lindat.mff.cuni.cz/services/udpipe/}} provide useful baselines, their performance can degrade in the presence of spelling variation, lexical sparsity, and genre-specific linguistic patterns. Recent advances in large language models (LLMs) offer new opportunities to address these challenges by leveraging contextual representations learned from large-scale multilingual corpora. However, systematic evaluation of LLM-based approaches for medieval Romance POS tagging remains limited, particularly with respect to prompting strategies, fine-tuning regimes, and multilingual transfer effects.

This study provides a comparative evaluation of traditional and neural methods across three medieval Romance datasets representing different genealogical branches and textual domains: Medieval Occitan, Medieval Catalan, and Medieval French. We investigate the following research questions:

\begin{enumerate}
    \item How do LLM-based approaches compare to established POS tagging tools for medieval Romance varieties?
    \item To what extent do prompting strategies and decoding configurations influence tagging performance?
    \item Can cross-lingual transfer learning improve performance across related historical languages?
\end{enumerate}

Our contributions are threefold. First, we present a controlled empirical comparison of traditional taggers and modern LLMs across multiple training and inference regimes. Second, we analyze the effectiveness of bilingual and trilingual transfer configurations, highlighting the role of linguistic proximity in multilingual historical NLP. Third, we provide practical recommendations for selecting modeling strategies under different resource constraints. Together, these contributions aim to support the development of more robust annotation pipelines for historical text collections.

\section{Related Work}

Computational processing of historical language has evolved from early rule-based approaches toward statistical and neural methods that better accommodate linguistic variation and limited supervision. Foundational surveys \citep{piotrowski2012natural} highlight key challenges in historical NLP, including orthographic instability, domain specificity, and data scarcity. Subsequent work has explored normalization and annotation strategies for early language stages, demonstrating the importance of adapting preprocessing and modeling techniques to non-standard linguistic input \citep{scheible2011token}.

For Romance historical varieties, research has focused on developing specialized corpora and task-specific annotation tools. Studies on Classical and Early Modern French show that domain-adapted models can substantially improve lemmatization and POS tagging accuracy compared to general-purpose systems \citep{camps2021corpus}. Neural approaches have further expanded these capabilities. Work on historical text normalization and lemmatization demonstrates that sequence-to-sequence and contextualized models can effectively capture variation-rich morphological patterns \citep{bollmann2019large, manjavacas2019improving, kestemont2016lemmatization}. Related efforts in OCR post-correction and handwritten text recognition highlight the broader applicability of neural pipelines in historical document processing \citep{springmann2016automatic, koch-etal-2023-tailored, arias-etal-2023-automatic, Wiedner2023, pavlopoulos-etal-2024-challenging, sarawgi2026digitizingnepalswrittenheritage}.

\noindent Recent studies have begun to investigate large language models in historical linguistic contexts. Prompt-based experiments suggest that pretrained multilingual models can transfer useful linguistic knowledge to low-resource historical varieties, although systematic comparisons across training regimes remain scarce \citep{schoeffel2025modernmodelsmedievaltexts, schoeffel2025unveiling}. In parallel, research on cross-lingual transfer learning indicates that multilingual training can improve performance on under-resourced languages when genealogical or typological proximity is present \citep{karthikeyan2020cross, muller2021first}. Building on this line of work, the present study provides a unified evaluation of traditional taggers, prompting-based LLM inference, fine-tuning, and multilingual transfer learning for medieval Romance POS tagging. By jointly analyzing multiple datasets and transfer configurations, we aim to clarify the relative strengths of these approaches in realistic historical NLP scenarios.

\section{Methodology}

We design a comparative experimental framework to evaluate traditional and neural approaches to part-of-speech (POS) tagging for medieval Romance languages. The study considers five experimental settings: direct application of traditional taggers, prompting-based large language model (LLM) inference, monolingual fine-tuning, bilingual cross-lingual transfer learning, and trilingual multilingual training. This setup enables systematic analysis of how varying levels of supervision and multilingual exposure affect tagging performance. In particular, the bilingual and trilingual configurations allow us to examine the role of linguistic relatedness and dataset composition in cross-lingual transfer. An overview of the experimental configuration is provided in Table~\ref{tab:experimental_setup}.

\subsection{Datasets}

We employ three historically grounded datasets representing distinct medieval Romance varieties and textual domains, summarized in Table~\ref{tab:datasets}. Further dataset references are available in the accompanying GitHub repository. The three resources differ substantially in size, genre, and century of composition, which we revisit when interpreting cross-lingual transfer results in Section~\ref{sec:results} and as a limitation in the closing discussion.
\newpage

\begin{table}[h]
\centering
\small
\begin{tabular}{@{}lllr@{}}
\toprule
\textbf{Dataset} & \textbf{Language} & \textbf{Genre} & \textbf{Tokens} \\
\midrule
NAF  & Med.\ Occitan (14th c.)  & Literary       & 45{,}457 \\
CAT      & Med.\ Catalan (13th c.)  & Chronicle      & 59{,}359 \\
Chauliac & Med.\ French (15th c.)   & Medical        &  2{,}443 \\
\bottomrule
\end{tabular}
\caption{Summary of the three medieval Romance datasets used in this study.}
\label{tab:datasets}
\end{table}

\textbf{Medieval Occitan}: the Nouvelle Acquisition Française 6195 (NAF6195), also known as manuscript M of the \textit{Vida de Sant Honorat}, dating from the 14th century. This literary manuscript reflects Provençal tradition and contains approximately 45,457 manually annotated tokens \citep{wiedner_2025_15300719}. \textbf{Medieval Catalan}: the \textit{Llibre dels Fets}, a historical chronicle describing the reign of James I of Aragon, composed in the 13th century. This chronicle comprises approximately 59,359 tokens with consistent morphological annotation \citep{pujol_i_campeny_2021_5615759}. \textbf{Medieval French}: the \textit{Anathomie} section from Gui de Chauliac’s \textit{Grande Chirurgie}, a 15th-century medical treatise containing specialized technical vocabulary and approximately 2,443 annotated tokens \citep{alma991083835439705501}\footnote{All datasets were preprocessed through tokenization and sentence segmentation, followed by manual verification of annotations. Tagsets were harmonized using Universal Dependencies conventions to ensure cross-dataset comparability.}.

\subsection{Models}

We include traditional POS taggers as baselines. \textbf{UDPipe} \citep{straka2016udpipe} is a neural pipeline trained on Universal Dependencies treebanks, offering tokenization, tagging, lemmatization, and parsing for many modern languages. \textbf{COLaF} is a neural tagger from the COLaF project covering French, the languages of France, and modern Occitan. Neither is natively trained on Medieval Catalan or Medieval Occitan, providing a realistic baseline for off-the-shelf generalization to historical Romance varieties. We additionally evaluate two open-source LLMs, Gemma3-12B \citep{team2024gemma} and Phi4-14B \citep{abdin2024phi}, under prompting, monolingual fine-tuning, and multilingual fine-tuning. Except for COLaF and UDPipe on Medieval French, none of the evaluated models were explicitly trained on Medieval Occitan, Catalan, or French. Representative language coverage per model is summarized in Table~\ref{tab:lang_support} (Appendix~\ref{a:supported_languages}).

\subsection{Prompting Strategies}

As summarized in Table~\ref{tab:prompting_strategies}, we investigate two prompting configurations: zero-shot and few-shot. In the \textbf{zero-shot} setting, models receive task instructions specifying the Universal Dependencies POS tagset and output format without exposure to target-domain examples. In the \textbf{few-shot} setting, prompts additionally include a small fixed set of annotated tokens reflecting morphological variability and orthographic diversity across the three target varieties. The example block is held constant across all evaluations to ensure prompt-level comparability and to avoid variance introduced by random example sampling. We deliberately mix examples from all three medieval varieties in a single block to expose the model to the orthographic heterogeneity it must cope with at inference time, rather than to optimize per-language prompts; we revisit this design choice as a limitation. This design allows assessment of how limited in-context supervision influences tagging performance.

\begin{table*}[h]
    \centering
    \small
    \begin{tabular}{|l|p{10cm}|}
        \hline
        \textbf{Prompting Strategy} & \textbf{Prompt} \\
        \hline
        Zero-shot & 
        \begin{minipage}[t]{10cm}
        \textit{You are a linguistic expert in Medieval Romance languages.\\[0.3em]
        Analyze the given text and assign Universal Dependencies Part-of-Speech tags (UPOS) to each token.\\[0.3em]
        Available tags: ``ADJ'', ``ADP'', ``ADV'', ``AUX'', ``CCONJ'', ``DET'', ``INTJ'', ``NOUN'', ``NUM'', ``PART'', ``PRON'', ``PROPN'', ``PUNCT'', ``SCONJ'', ``VERB'', ``X'', ``SYM''.\\[0.3em]
        Return a JSON array of objects, each with only ``word'' and ``UPOS'' keys.\\[0.3em]
        Output only the JSON array, properly formatted and closed, with no extra text or explanation.}
        \end{minipage} \\
        \hline
        Few-shot & 
        \begin{minipage}[t]{10cm}
        \textit{\textbf{\textit{Zero-shot prompt}} + \\Consider syntactic and semantic relationships, including agreement, word order, and morphology. Medieval Romance languages often exhibit significant spelling variation; for example, Old Occitan: `ansy', `eynsi', or `anes'; Old Catalan: `fiyl', or `conseyl'; Middle French: `norryr' or `norrir'.\\[0.5em]
        Example format:}\\[0.3em]
        \texttt{\footnotesize\lbrack\{``word'': ``bo'', ``UPOS'': ``ADJ''\}, \{``word'': ``volch'', ``UPOS'': ``VERB''\}, \{``word'': ``seyor'', ``UPOS'': ``NOUN''\}, \{``word'': ``homps'', ``UPOS'': ``NOUN''\}, \{``word'': ``sant'', ``UPOS'': ``ADJ''\}, \{``word'': ``iorn'', ``UPOS'': ``NOUN''\}, \{``word'': ``ilz'', ``UPOS'': ``PRON''\}, \{``word'': ``addicions'', ``UPOS'': ``NOUN''\}, \{``word'': ``deffendre'', ``UPOS'': ``VERB''\}\rbrack}
        \end{minipage} \\
        \hline
    \end{tabular}
    \caption{Comparison of different prompting strategies for UD POS tagging.}
    \label{tab:prompting_strategies}
\end{table*}

\subsection{Decoding Strategies}

To assess the robustness of prompting-based inference, we evaluate several commonly used decoding approaches \citep{wiher-etal-2022-decoding,garces-arias-etal-2025-decoding}. Specifically, we consider beam search, temperature sampling \citep{ackley1985learning}, top-$k$ sampling \citep{fan-etal-2018-hierarchical}, and nucleus (top-$p$) sampling \citep{holtzman2019curious}. Hyperparameter ranges for each decoding configuration are summarized in Table~\ref{tab:experimental_setup}.

\subsection{Fine-tuning Experiments}

We investigate three fine-tuning scenarios, all using a single fixed 80\%/20\% train/test split per dataset. The same 20\% test partition for each dataset is reused across the monolingual, bilingual, and trilingual settings, so that test tokens are never seen during training in any configuration and accuracy numbers are directly comparable across regimes.

First, in \textbf{monolingual fine-tuning}, each LLM is trained on the 80\% partition of one dataset and evaluated on the corresponding 20\% test partition.

Second, in \textbf{bilingual cross-lingual transfer learning} (CLTF), models are trained on the union of the 80\% partitions of two datasets (CAT+OCC, CAT+FR, FR+OCC, where OCC = NAF and FR = Chauliac) and evaluated on the held-out 20\% partition of each constituent language.
\newpage

\noindent This setup allows analysis of how genealogical proximity and corpus characteristics influence transfer effectiveness. Third, in \textbf{trilingual CLTF}, models are trained on the combined 80\% partitions of all three datasets and evaluated separately on each language's 20\% test partition. This configuration tests whether broader multilingual exposure improves performance, particularly for lower-resource varieties. A true held-out cross-lingual evaluation (e.g., CAT+FR$\rightarrow$NAF, with the target language entirely absent from training) is left for future work. Fine-tuning hyperparameters are provided in Table~\ref{tab:finetuning_hyperparams}.

\subsection{Evaluation Metrics}

Model performance is assessed using standard classification metrics (see Appendix~\ref{a:metrics}). \textbf{Accuracy} measures the proportion of correctly predicted POS tags across all tokens, providing an overall performance indicator. \textbf{Macro-averaged F1} captures balanced performance across frequent and infrequent POS categories by averaging class-level F1-scores.

\subsection{Experimental Setup}

A comprehensive description of the models, datasets, experiments and tasks is summarized in Table \ref{tab:experimental_setup}.

\begin{table*}[h]
\centering
\small
\renewcommand{\arraystretch}{1.2}
\definecolor{headercolor}{RGB}{240,240,240}
\begin{tabular}{p{4cm}p{9cm}}
\toprule
\rowcolor{headercolor} \multicolumn{2}{c}{\textbf{Models \& Datasets}} \\
\midrule
\textbf{Traditional} & COLaF, UDPipe\\
\textbf{LLMs} & Gemma3-12B \citep{team2024gemma}, Phi4-14B \citep{abdin2024phi}\\
& \textit{Language support in Table \ref{tab:lang_support}, Appendix \ref{a:supported_languages}}\\
\textbf{Datasets} & NAF (Medieval Occitan, 14th c.), CAT (Medieval Catalan, 13th c.), Chauliac (Medieval French, 15th c.)\\
\midrule
\rowcolor{headercolor} \multicolumn{2}{c}{\textbf{Experimental Tasks}} \\
\midrule
\textbf{Task 1: Traditional} & Direct evaluation using COLaF and UDPipe on all datasets\\
\midrule
\textbf{Task 2: LLM Prompting} & 
Zero-shot \& few-shot prompting (Table \ref{tab:prompting_strategies})\\
& Decoding: beam search ($w \in \{1,15\}$), temperature ($\tau \in \{0.6,0.8,0.9\}$), top-$k$ ($k \in \{5,20,50\}$), top-$p$ ($p \in \{0.75,0.85,0.95\}$)\\
\midrule
\textbf{Task 3: LLM Fine-tuning} & 
80/20 train/test split per dataset\\
& Each model fine-tuned and tested on same dataset (1-to-1 mapping)\\
\midrule
\textbf{Task 4: Bilingual CLTF} &
80\% of two datasets for training, 20\% per dataset for testing\\
& Pairwise: CAT+OCC, CAT+FR, FR+OCC (2-to-1 transfer)\\
\midrule
\textbf{Task 5: Trilingual CLTF} & 
80\% of all datasets for training, 20\% per dataset for testing\\
& Multilingual training $\rightarrow$ monolingual testing (N-to-1 transfer)\\
\bottomrule
\end{tabular}
\caption{Experimental setup for POS tagging of medieval Romance languages. Evaluation focused on accuracy with precision, recall, and F1-measures available (Appendix \ref{a:metrics}). All experiments used NVIDIA H100-96GB GPU. Hyperparameters detailed in Appendices \ref{a:llm_prompting_hyperparams} and \ref{a:finetuning_hyperparams}.}
\label{tab:experimental_setup}
\end{table*}

\section{Results}
\label{sec:results}

\subsection{Overall Performance Comparison}

\begin{figure}[h]
    \centering
    \includegraphics[width=1.03\columnwidth]{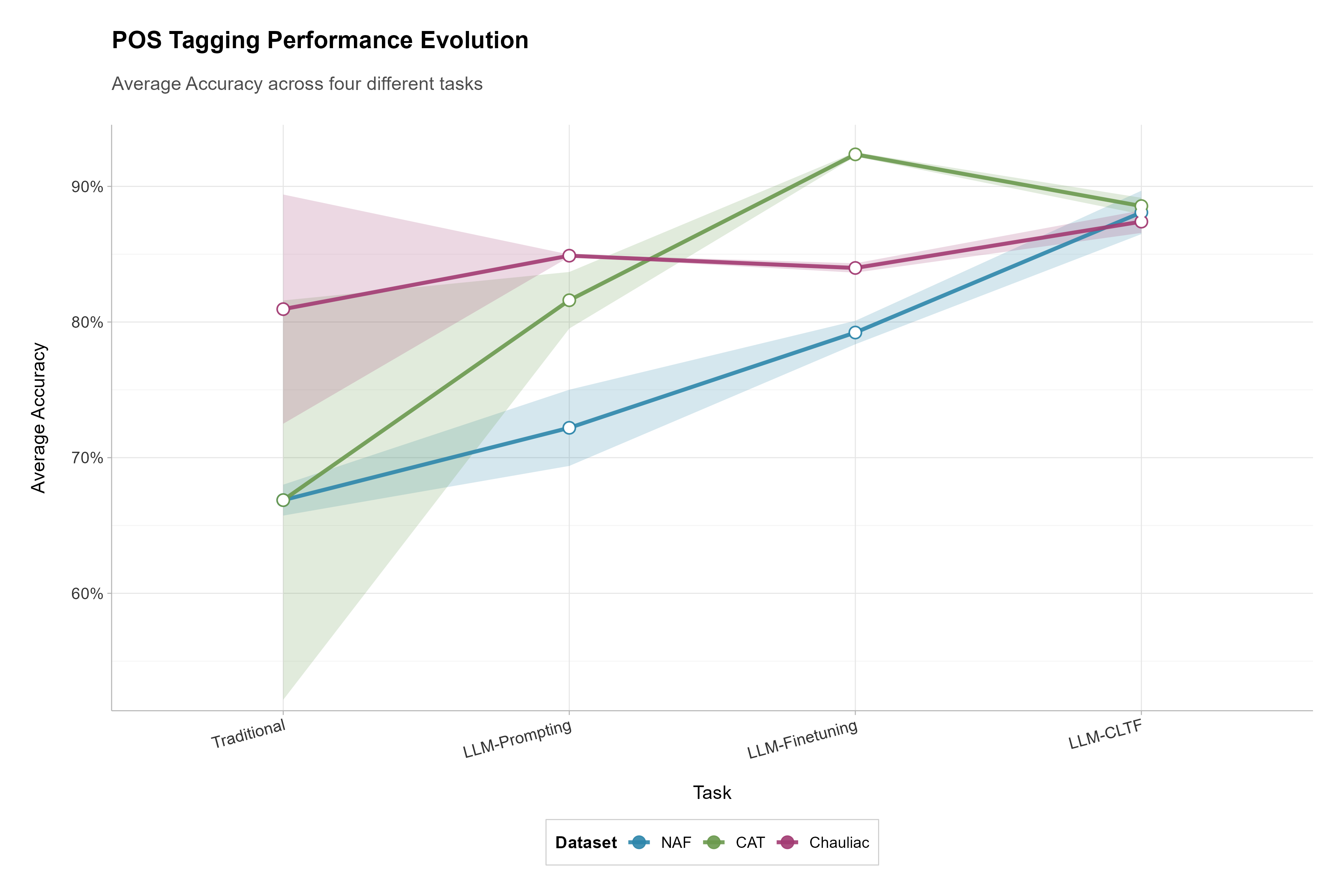}
\caption{Illustration of performance evolution, in terms of Average Accuracy, per dataset (NAF in blue, CAT in green, Chauliac in purple), from traditional POS taggers (UDPipe, COLaF) through LLM prompting and monolingual fine-tuning to trilingual cross-lingual transfer learning (CLTF). Shaded areas represent variability across model and decoding configurations within each method family.}
    \label{fig:method_evolution}
\end{figure}

\newpage
Figure~\ref{fig:method_evolution} summarizes tagging accuracy across experimental configurations. Traditional POS taggers achieve competitive results on selected datasets but exhibit substantial performance variability across language varieties. Prompting-based LLM inference generally yields higher and more stable accuracy, while supervised fine-tuning further improves performance across all datasets. The largest gains are observed under multilingual training regimes, particularly for the lower-resource Medieval Occitan dataset.

In aggregate terms, traditional approaches reach an average accuracy of 71.56\%, with pronounced dispersion across cells (52.15\%--89.40\%). Prompting-based LLM inference improves average accuracy to 77.75\% (range 62.53\%--84.98\% across cells), while also reducing per-dataset variability relative to the traditional baselines. Dataset-specific patterns are evident: the Medieval Catalan dataset exhibits the largest performance gap between traditional systems and fine-tuned LLMs (10.93 percentage points), whereas improvements for NAF and Chauliac are more moderate (approximately 5 percentage points on average).

Monolingual LLM fine-tuning yields a further increase in mean accuracy to 85.19\% (range 78.36\%--92.52\% across cells). The strongest results are obtained on the CAT dataset, where both Gemma3 and Phi4 exceed 92\% accuracy. At the same time, the broader performance range observed under fine-tuning suggests sensitivity to dataset size, domain, and linguistic characteristics.

Bilingual cross-lingual transfer learning (CLTF) produces heterogeneous effects depending on language pairing. The CAT+OCC configuration achieves 89.25\% accuracy on NAF, approaching trilingual performance levels, while CAT+FR reaches 93.14\% on Chauliac, representing the highest accuracy observed for Medieval French across all experimental conditions. This bilingual configuration outperforms both monolingual fine-tuning (+9.50 percentage points) and trilingual CLTF (+4.91 percentage points) for this dataset. In contrast, the FR+OCC pairing results in only marginal improvement on NAF (80.31\% vs.\ 80.09\% monolingual), indicating that transfer effectiveness depends strongly on the specific language combination.

Trilingual CLTF achieves an average accuracy of 88.01\% with the narrowest spread among all method families (range 86.48\%--89.68\% across cells), suggesting comparatively stable behavior across datasets. Relative to the UDPipe baseline, trilingual training leads to substantial gains on NAF (+21.67 percentage points) and CAT (+7.57 percentage points), while performance on Chauliac decreases slightly ($-1.17$ percentage points). Notably, the bilingual CAT+FR configuration remains superior for Medieval French, illustrating that broader multilingual exposure does not necessarily yield optimal results for all target languages. A detailed comparison is provided in Table~\ref{tab:overall_performance}\footnote{\textbf{Caveat.} Token counts differ by more than an order of magnitude across datasets (Table~\ref{tab:datasets}), so the strong CAT+FR$\rightarrow$Chauliac result partly reflects the additional training material contributed by Catalan rather than genealogical proximity alone. Disentangling size, genre, and relatedness through controlled subsampling is left for future work.}.

Overall, the progression from traditional methods (71.56\%) to prompting (77.75\%), fine-tuning (85.19\%), and multilingual transfer (up to 88.01\%) reflects systematic improvements associated with increased supervision and multilingual training signals. However, the results also suggest that optimal transfer configurations are language-dependent: trilingual training appears beneficial for Medieval Occitan, whereas targeted bilingual pairing with Catalan is more effective for Medieval French.

\begin{table*}
\centering
\resizebox{0.8\textwidth}{!}{%
\begin{tabular}{@{}llccccccc@{}}
\toprule
\multirow{2}{*}{\textbf{Task}} & \multirow{2}{*}{\textbf{Model/Strategy}} & \multicolumn{2}{c}{\textbf{NAF}} & \multicolumn{2}{c}{\textbf{CAT}} & \multicolumn{2}{c}{\textbf{Chauliac}} \\
\cmidrule(lr){3-4} \cmidrule(lr){5-6} \cmidrule(lr){7-8}
& & \textbf{Acc.} & \textbf{F1} & \textbf{Acc.} & \textbf{F1} & \textbf{Acc.} & \textbf{F1} \\
\midrule
\multirow{2}{*}{Traditional} & UDPipe & \textbf{68.01} & \textbf{67.29} & \textbf{81.59} & \textbf{81.19} & \textbf{89.40} & \textbf{89.53} \\
& COLaF & 65.73 & 65.47 & 52.15 & 51.50 & 72.50 & 67.43 \\
\midrule
\multirow{4}{*}{Prompting} & Gemma3 Zero-shot & 62.53 & 61.81 & 72.54 & 74.03 & 82.49 & 82.58 \\
& Gemma3 Few-shot & 69.39 & 69.22 & 79.48 & 80.49 & 84.80 & 85.20 \\
& Phi4 Zero-shot & 72.78 & 71.94 & 80.84 & 81.01 & 84.45 & 84.61 \\
& Phi4 Few-shot & \textbf{75.01} & \textbf{74.31} & \textbf{83.69} & \textbf{83.75} & \textbf{84.98} & \textbf{85.19} \\
\midrule
\multirow{2}{*}{Fine-tuning} & Gemma3 & \textbf{80.09} & \textbf{79.99} & \cellcolor{bestresult}\textbf{92.52} & \cellcolor{bestresult}\textbf{92.50} & 83.64 & 83.74 \\
& Phi4 & 78.36 & 78.35 & 92.20 & 92.13 & \textbf{84.33} & \textbf{84.10} \\
\midrule
\multirow{3}{*}{Bilingual CLTF} 
& Gemma3 (CAT+OCC) & \textbf{89.25} & \textbf{89.18} & 91.62 & 91.54 & -- & -- \\
& Gemma3 (CAT+FR)  & -- & -- & 91.28 & 91.19 & \cellcolor{bestresult}\textbf{93.14} & \cellcolor{bestresult}\textbf{93.02} \\
& Gemma3 (FR+OCC)  & 80.31 & 80.22 & -- & -- & 85.74 & 85.61 \\
\midrule
\multirow{2}{*}{Trilingual CLTF} & Gemma3 & \cellcolor{bestresult}\textbf{89.68} & \cellcolor{bestresult}\textbf{89.66} & \textbf{89.16} & \textbf{89.11} & \textbf{88.23} & \textbf{88.09} \\
& Phi4 & 86.48 & 86.39 & 87.94 & 87.74 & 86.57 & 86.48 \\
\midrule
\midrule
$\Delta_\text{best, UDPipe}$ & Best CLTF vs UDPipe & +21.67 & +22.37 & +7.57 & +7.92 & +3.74 & +3.49 \\
\bottomrule
\end{tabular}%
}
\caption{Overall performance comparison across methods and datasets. Best result per method is in \textbf{bold}; best overall result per column is highlighted in \colorbox[HTML]{D4EDDA}{\textbf{green}}. For each bilingual CLTF row, results are reported on the held-out partition of each constituent language; cells marked ``--'' indicate languages outside the training pair.}
\label{tab:overall_performance}
\end{table*}

\subsection{Task-Specific Analysis}

\subsubsection{Traditional vs. LLM-based Approaches}

Across datasets, LLM-based methods consistently outperform traditional POS taggers. UDPipe achieves relatively strong results on the Chauliac dataset (89.40\% accuracy), but performance declines substantially on NAF (68.01\%), highlighting the challenges posed by orthographic variability and the limited training data available in Medieval Occitan.

\subsubsection{Prompting Strategy Effectiveness}

Few-shot prompting yields systematic improvements over zero-shot configurations across all models and datasets (see Figure~\ref{fig:decoding_performance} in Appendix~\ref{a:decoding}). Absolute gains range from 2.94 percentage points on Chauliac with Gemma3 to 10.24 percentage points on NAF with Phi4. Among the evaluated models, Phi4 demonstrates stronger prompting performance overall. In addition, deterministic decoding strategies tend to outperform sampling-based approaches, with beam search (beam width = 15) consistently achieving the highest accuracy. Detailed decoding-level results and variability analyses are provided in Tables~\ref{tab:decoding_comprehensive} and~\ref{tab:decoding_robustness} (Appendix~\ref{a:decoding_effects}).

\subsubsection{Monolingual Fine-tuning vs.\ Multilingual CLTF}

\begin{figure*}[htb!]
    \centering
    \includegraphics[width=0.9\textwidth]{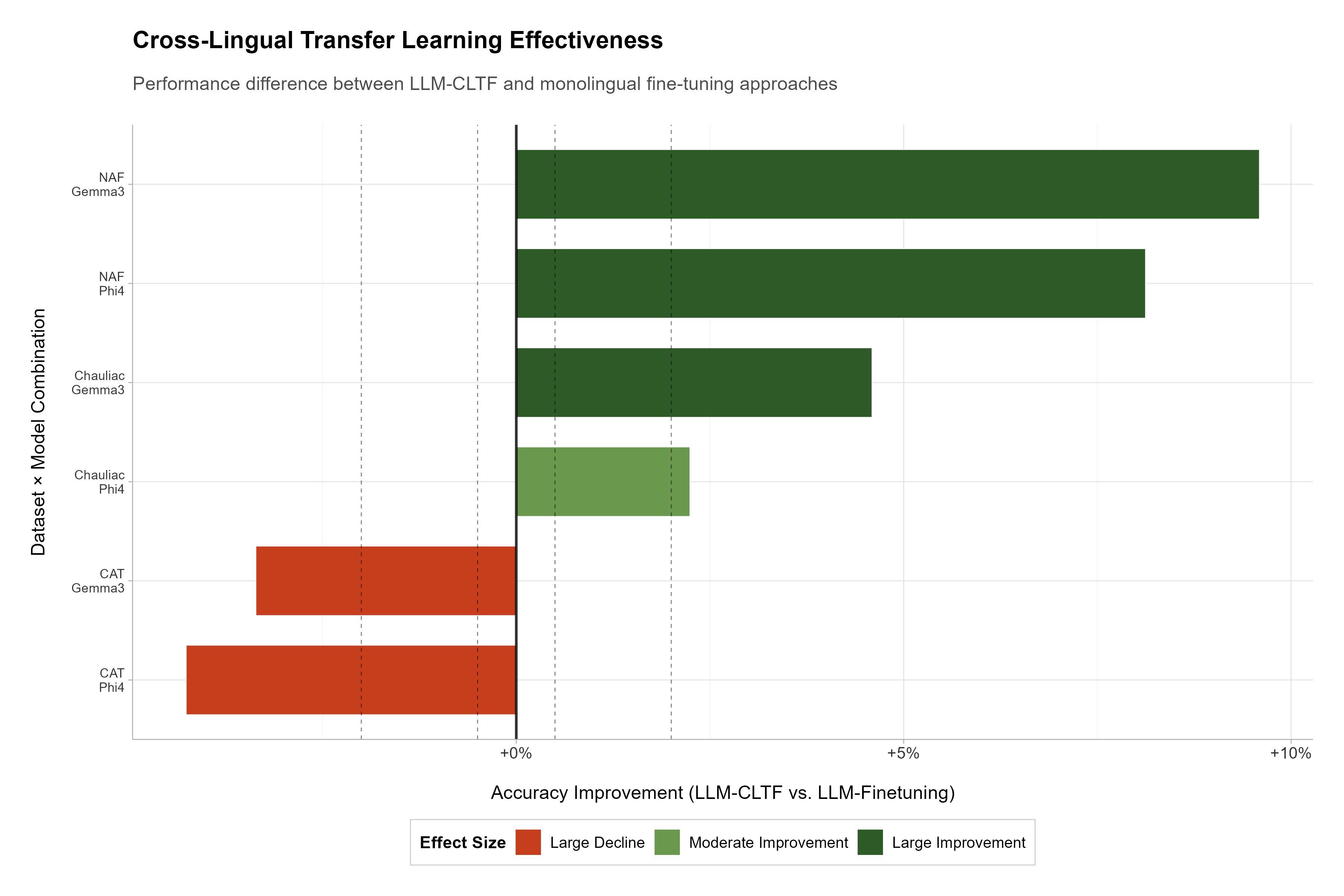}
    \caption{Effect of trilingual CLTF with respect to single-dataset LLM fine-tuning.}
    \label{fig:cltf_impact}
\end{figure*}

Monolingual fine-tuning achieves the highest performance on Medieval Catalan (92.52\% with Gemma3). In contrast, trilingual CLTF yields the largest improvement for Medieval Occitan, increasing accuracy from 80.09\% under monolingual fine-tuning to 89.68\%. This pattern suggests that cross-lingual supervision is particularly beneficial for lower-resource datasets. The relative impact of multilingual training across datasets and models is illustrated in Figure~\ref{fig:cltf_impact}.

\subsubsection{Bilingual vs. Trilingual Transfer}

Bilingual CLTF results indicate that targeted language pairing can match or exceed trilingual training for specific datasets (see Table~\ref{tab:overall_performance}). Three observations emerge. First, \textbf{Medieval Catalan functions as an effective bridge language}. Pairing Catalan with either Occitan (CAT+OCC $\rightarrow$ NAF: 89.25\%) or French (CAT+FR $\rightarrow$ Chauliac: 93.14\%) leads to substantial gains over monolingual fine-tuning. This pattern is consistent with Catalan’s intermediate genealogical position between Gallo-Romance and Occitano-Romance varieties. Second, \textbf{transfer effectiveness varies across language combinations}.

\noindent The configuration FR+OCC yields only minimal improvement for NAF (+0.22 percentage points), suggesting that linguistic proximity alone does not guarantee successful transfer. Corpus size imbalance and domain differences may also influence performance.

\begin{table*}[htb!]
\centering
\small
\setlength{\tabcolsep}{4pt}
\begin{tabular}{@{}lccccc@{}}
\toprule
\textbf{POS} & \textbf{UDPipe} & \textbf{Phi4 FS} & \textbf{Gemma3 FT} & \textbf{Gemma3 CLTF} & $\boldsymbol{\Delta}$ \\
\midrule
PROPN & 25.85 & 72.34 & 78.31 & \cellcolor{bestresult}\textbf{92.47} & +66.62 \\
NUM   & 28.92 & 61.39 & \cellcolor{bestresult}\textbf{91.89} & 86.01 & +62.97 \\
AUX   & 38.39 & 45.08 & 53.58 & \cellcolor{bestresult}\textbf{61.04} & +22.65 \\
PRON  & 45.80 & 52.77 & 76.38 & \cellcolor{bestresult}\textbf{81.51} & +35.71 \\
ADV   & 50.61 & 54.19 & 66.92 & \cellcolor{bestresult}\textbf{74.38} & +23.77 \\
SCONJ & 52.94 & 54.73 & 57.97 & \cellcolor{bestresult}\textbf{94.62} & +41.68 \\
ADJ   & 65.29 & 72.05 & 71.17 & \cellcolor{bestresult}\textbf{73.58} &  +8.29 \\
VERB  & 67.77 & 79.91 & 75.79 & \cellcolor{bestresult}\textbf{89.00} & +21.23 \\
\midrule
DET   & 73.81 & 73.48 & \cellcolor{bestresult}\textbf{89.97} & 87.99 & +16.16 \\
NOUN  & 76.45 & 83.65 & 82.81 & \cellcolor{bestresult}\textbf{89.44} & +12.99 \\
CCONJ & 83.06 & 81.72 & 86.67 & \cellcolor{bestresult}\textbf{96.34} & +13.28 \\
ADP   & 85.51 & 89.93 & 88.59 & \cellcolor{bestresult}\textbf{92.78} &  +7.27 \\
\bottomrule
\end{tabular}
\caption{Per-class F1 on \textbf{NAF}: low-performing (top) and high-performing (bottom) classes. FS = few-shot, FT = fine-tuned, $\Delta$ = improvement of best LLM-based method over UDPipe. Best per row in \colorbox[HTML]{D4EDDA}{\textbf{green}}.}
\label{tab:pos_performance_naf}
\end{table*}

Third, \textbf{increasing the number of training languages does not uniformly improve results}. For Medieval French, bilingual CAT+FR training (93.14\%) clearly outperforms trilingual CLTF (88.23\%). This likely reflects the small size of the Chauliac dataset, which benefits from focused exposure to closely related Catalan material; all three corpora are domain-specific, so corpus size appears to be the more salient factor here. Conversely, Medieval Occitan shows the strongest performance under trilingual training, suggesting complementary transfer signals from both Catalan and French.

\section{Error Analysis}
\label{sec:error_analysis}

\subsection{Part-of-Speech Class Performance}

Analysis of POS-specific F1-scores reveals systematic differences across methods. Table~\ref{tab:pos_performance_naf} (NAF) is shown here; CAT and Chauliac results are in Appendix~\ref{a:pos_class_effect}. Adjectives and adverbs remain particularly challenging for traditional approaches, with UDPipe yielding comparatively low F1-scores for these categories. LLM-based approaches, especially fine-tuned models, show marked improvements, which suggests that contextual representations may be helpful for disambiguating semantically richer POS classes. Pronouns also show consistent gains across LLM-based methods, potentially reflecting improved handling of context-dependent dependencies.

By contrast, function words such as adpositions and coordinating conjunctions maintain high performance across most methods. This pattern suggests that these categories may be more reliably identified from local syntactic cues. Verb classification also remains comparatively stable across approaches, indicating that morphological marking provides relatively strong signals even in the presence of orthographic variation.

\subsection{Cross-Lingual Transfer Effects}

Cross-lingual transfer results indicate that content-word categories (e.g., nouns, adjectives, verbs) tend to benefit more from multilingual training than function-word categories. The NAF dataset shows the largest gains under trilingual CLTF, with accuracy increasing from 80.09\% under monolingual fine-tuning to 89.68\%. These improvements are particularly pronounced for lower-frequency POS classes, suggesting that multilingual exposure can help mitigate data sparsity in historical language processing. The bilingual transfer results refine this picture further. The strong performance of the CAT+OCC configuration on NAF suggests that Catalan provides useful lexical and morphological information for Occitan. Similarly, the gains observed for CAT+FR on Chauliac indicate that focused bilingual training may be advantageous for small, domain-specific datasets.

\section{Discussion: Implications for Historical NLP}

The results of this study provide several insights for the development of computational pipelines in historical language processing. First, the observed performance differences across training regimes suggest that model adaptation strategies should be carefully aligned with resource availability and linguistic relatedness. In particular, multilingual training appears to offer robust benefits for lower-resource historical varieties, while targeted bilingual configurations may yield stronger performance for small, domain-specific datasets.

Second, the analysis highlights the importance of contextual modeling for linguistically variable input. Improvements observed for semantically richer POS classes such as adjectives, adverbs, and pronouns indicate that contextualized neural representations can mitigate challenges associated with orthographic instability and morphological variation. This finding supports recent trends toward integrating large pretrained models into digital humanities workflows, where annotation robustness is often constrained by heterogeneous textual sources.

Third, the heterogeneous transfer effects observed across language pairings underline the need for historically informed model design. Genealogical proximity alone does not fully determine transfer success; corpus size, genre, and annotation consistency also appear to play important roles. These factors suggest that future historical NLP research should move beyond purely language-centric transfer assumptions and adopt more data-driven criteria for multilingual training configuration. These findings indicate that modern neural tagging approaches can meaningfully complement traditional linguistic annotation methods. 

\section{Conclusion}
\label{sec:conclusion}

This paper presents a systematic empirical evaluation of large language models for POS tagging across three medieval Romance language datasets. Relative to traditional tagging tools, LLM-based approaches yield consistent performance improvements under prompting, fine-tuning, and multilingual training settings. Cross-lingual transfer learning is particularly beneficial for lower-resource varieties, while targeted bilingual configurations can outperform broader multilingual training for specific target languages. Beyond reporting performance differences, the analysis highlights the importance of linguistic proximity and dataset composition when designing multilingual training strategies for historical NLP. Overall, the findings suggest that contemporary neural models can contribute to more reliable linguistic annotation pipelines for medieval texts and may support broader digital humanities workflows. Future work may include additional historical languages, examine downstream task effects, and explore alternative multilingual adaptation strategies.

\paragraph{Limitations}
Our study has several limitations. First, dataset sizes range from 2{,}443 to 59{,}359 tokens and represent different genres. We did not run controlled subsample or learning-curve experiments, so corpus size, genre, and genealogical proximity cannot be fully disentangled in the cross-lingual transfer results. Second, the few-shot block mixes examples from all three varieties; per-language prompts and example-selection strategies were not systematically explored. Third, the strongest results are obtained with fine-tuned 12--14B LLMs, which are substantially more expensive than UDPipe or COLaF. We did not evaluate smaller LLMs (e.g., 1--3B) or a modern pre-trained encoder fine-tuned for token classification (e.g., XLM-R). Moreover, Gemma3-12B and Phi4-14B were chosen as strong openly available instruction-tuned models that fit on a single H100 under LoRA; the 7--9B range and proprietary models were not explored. Finally, generalization to other historical language families and to downstream tasks (lemmatization, parsing, NER) remains untested.

\section*{Ethics Statement}

We affirm that our research adheres to the \href{https://www.aclweb.org/portal/content/acl-code-ethics}{ACL Ethics Policy}. This work uses publicly available datasets and involves no human subjects or personally identifiable information. All code and data are released to enable reproducible research and further investigation.

\section*{Acknowledgments}
We would like to express our gratitude to the ALMA (\textit{Wissensnetze in der mittelalterlichen Romania}) project for their support and for granting us access to a partially annotated subset of the Chauliac dataset. We also sincerely thank Marinus Wiedner for his work on the annotation and publication of Medieval Occitan corpora. His efforts provided an important resource that made the broader analyses in the present work possible. Additionally, we thank the Leibniz-Rechenzentrum der Bayerischen Akademie der Wissenschaften (LRZ) for providing the computational resources essential to this research. Esteban Garces Arias sincerely thanks the Mentoring Program of the Faculty of Mathematics, Statistics, and Informatics at LMU Munich and the Munich Center for Machine Learning (MCML) for their ongoing mentorship and financial support.

\bibliography{custom}

\clearpage

\onecolumn
\appendix
\section*{Appendix}

\section{Supported languages (pre-training)}
\label{a:supported_languages}

\begin{table}[H]
\centering
\rowcolors{1}{}{white} 
\resizebox{0.80\textwidth}{!}{%
\begin{tabular}{lcccc}
\hline
\textbf{Language}            & \textbf{COLaF} & \textbf{UDPipe} & \textbf{Phi4-14B} & \textbf{Gemma3-12B} \\
\hline
Occitan (modern)              & \cmark         &           & \cmark            &                    \\
\rowcolor{gray!15}
\textbf{Medieval Occitan}     &          &           &                   &                    \\
Catalan (modern)              &          & \cmark          & \cmark            &                    \\
\rowcolor{gray!15}
\textbf{Medieval Catalan}     &          &          &                   &                    \\
French (modern)               &     \cmark     & \cmark          & \cmark            & \cmark             \\
\rowcolor{gray!15}
\textbf{Medieval French}      & \cmark         & \cmark          &                   &                    \\
Spanish (modern)              &          & \cmark          & \cmark            & \cmark             \\
Italian (modern)              &          & \cmark          & \cmark            & \cmark             \\
Portuguese (modern)           &          & \cmark          & \cmark            & \cmark             \\
Romanian (modern)             &          & \cmark          & \cmark            & \cmark             \\
Galician (modern)             &          & \cmark          & \cmark            &                    \\
Asturian (modern)             &          &           & \cmark            &                    \\
Sardinian (modern)            &          &           & \cmark            &                    \\
Sicilian (modern)             &          &           & \cmark            &                    \\
Ligurian (modern)             &          &           & \cmark            &                    \\
Lombard (modern)              &          &           & \cmark            &                    \\
Venetian (modern)             &          &           & \cmark            &                    \\
Friulian (modern)             &          &           & \cmark            &                    \\
Arabic                        &          & \cmark          & \cmark            & \cmark             \\
English                       &          & \cmark          & \cmark            & \cmark             \\
\hline
\end{tabular}
}
\caption{Language support (modern vs.\ medieval) across traditional POS taggers (COLaF, UDPipe) and LLMs (Phi4-14B and Gemma3-12B).}
\label{tab:lang_support}
\end{table}

\section{Hyperparameters for LLM Prompting}
\label{a:llm_prompting_hyperparams}

\begin{table}[H]
\centering
\begin{tabular}{@{}lll@{}}
\toprule
\textbf{Category} & \textbf{Hyperparameter} & \textbf{Value} \\
\midrule
\multirow{3}{*}{Tokenizer} & Max Length & 8192 \\
                           & Padding Side & left \\
                           & Data Type & \makecell[l]{torch.bfloat16 (Gemma) \\ torch.float16 (Phi-4)} \\
\midrule
\multirow{2}{*}{Model} & Max New Tokens & 300 \\
                       & Batch Size & 8 \\
\midrule
\multirow{2}{*}{Processing} & Chunk Size & 20 \\
                           & Window Length & 5 \\
\bottomrule
\end{tabular}
\caption{Hyperparameters used for LLM prompting experiments.}
\label{tab:llm_prompting_hyperparams}
\end{table}

\section{Hyperparameters for LLM Fine-tuning}
\label{a:finetuning_hyperparams}

\begin{table}[H]
\centering
\begin{tabular}{@{}lll@{}}
\toprule
\textbf{Category} & \textbf{Hyperparameter} & \textbf{Value} \\
\midrule
\multirow{4}{*}{LoRA} & LoRA Rank ($r$) & 16 \\
                      & LoRA Alpha ($\alpha$) & 32 \\
                      & LoRA Dropout & 0.1 \\
                      & Target Modules & \makecell[l]{q\_proj, v\_proj, \\ k\_proj, o\_proj} \\
\midrule
\multirow{5}{*}{Training} & Learning Rate & $2 \times 10^{-4}$ \\
                          & Batch Size & 4 \\
                          & Number of Epochs & 10 \\
                          & Optimizer & AdamW \\
                          & Weight Decay & 0.01 \\
\bottomrule
\end{tabular}
\caption{Hyperparameters used for LLM fine-tuning experiments with LoRA.}
\label{tab:finetuning_hyperparams}
\end{table}

\section{Performance Analysis}
\label{a:performance_analysis}

\subsection{Effect of Decoding Strategies}
\label{a:decoding_effects}

\begin{table}[H]
\centering
\resizebox{0.8\textwidth}{!}{%
\begin{tabular}{@{}llccccc@{}}
\toprule
\multirow{1}{*}{\textbf{Model}} & \multirow{1}{*}{\textbf{Strategy}} & \multicolumn{1}{c}{\textbf{NAF}} & \multicolumn{1}{c}{\textbf{CAT}} & \multicolumn{1}{c}{\textbf{Chauliac}} & \multicolumn{1}{c}{\textbf{Average}} & \multicolumn{1}{c}{\textbf{Std Dev}} \\
\midrule
\multirow{12}{*}{\textbf{Gemma3}} 
& Zero-shot + Beam-15 & 62.53 & 72.54 & 82.36 & 72.48 & 10.08 \\
& Few-shot + Beam-1 & 69.24 & 79.37 & 84.27 & 77.63 & 7.51 \\
& Few-shot + Beam-15 & \textbf{69.39} & \textbf{79.52} & 84.51 & 77.81 & 7.50 \\
& Few-shot + Top-$k$-5 & 69.22 & 79.48 & \textbf{84.80} & \textbf{77.83} & 7.79 \\
& Few-shot + Top-$k$-20 & 69.29 & 79.33 & 84.35 & 77.66 & 7.51 \\
& Few-shot + Top-$k$-50 & 69.17 & 79.41 & 84.28 & 77.62 & 7.56 \\
& Few-shot + Top-$p$-0.75 & 69.33 & 79.47 & 84.56 & 77.79 & 7.62 \\
& Few-shot + Top-$p$-0.85 & 69.34 & 79.42 & 84.44 & 77.73 & 7.54 \\
& Few-shot + Top-$p$-0.95 & 69.27 & 79.31 & 83.99 & 77.52 & \textbf{7.34} \\
& Few-shot + Temp-0.6 & 69.23 & 79.46 & 84.49 & 77.73 & 7.63 \\
& Few-shot + Temp-0.8 & 69.30 & 79.35 & 84.31 & 77.65 & 7.48 \\
& Few-shot + Temp-0.9 & 69.35 & 79.43 & 84.39 & 77.72 & 7.52 \\
\addlinespace
\multirow{12}{*}{\textbf{Phi4}}
& Zero-shot + Beam-15 & 72.77 & 80.84 & 84.09 & 79.23 & 6.67 \\
& Few-shot + Beam-1 & 74.86 & 83.47 & 84.60 & 80.98 & 5.37 \\
& Few-shot + Beam-15 & \cellcolor{bestresult}\textbf{75.01} & \cellcolor{bestresult}\textbf{83.69} & \cellcolor{bestresult}\textbf{84.98} & \cellcolor{bestresult}\textbf{81.23} & 5.32 \\
& Few-shot + Top-$k$-5 & 74.02 & 82.88 & 84.31 & 80.40 & 5.15 \\
& Few-shot + Top-$k$-20 & 73.76 & 82.85 & 83.98 & 80.20 & 5.55 \\
& Few-shot + Top-$k$-50 & 73.80 & 82.81 & 84.11 & 80.24 & 5.21 \\
& Few-shot + Top-$p$-0.75 & 74.50 & 83.46 & 84.51 & 80.82 & 5.53 \\
& Few-shot + Top-$p$-0.85 & 74.49 & 83.34 & 84.00 & 80.61 & 5.43 \\
& Few-shot + Top-$p$-0.95 & 74.19 & 83.16 & 84.56 & 80.64 & 5.70 \\
& Few-shot + Temp-0.6 & 74.48 & 83.23 & 84.53 & 80.75 & 5.53 \\
& Few-shot + Temp-0.8 & 74.27 & 82.94 & 83.78 & 80.33 & \cellcolor{bestresult}\textbf{4.84} \\
& Few-shot + Temp-0.9 & 74.26 & 82.97 & 84.69 & 80.64 & 5.95 \\
\bottomrule
\end{tabular}%
}
\caption{Comprehensive decoding strategy performance analysis. Best results per model are highlighted in \textbf{bold}, while best overall results per column are highlighted in \colorbox[HTML]{D4EDDA}{\textbf{green}}.}
\label{tab:decoding_comprehensive}
\end{table}

\begin{table}[H]
\centering
\begin{tabular}{@{}lccccc@{}}
\toprule
\textbf{Strategy Type} & \multicolumn{1}{c}{\textbf{Mean Acc.}} & \multicolumn{1}{c}{\textbf{Std Dev.}} & \multicolumn{1}{c}{\textbf{CV}} & \multicolumn{1}{c}{\textbf{Range}} & \textbf{Recommendation} \\
\midrule
\textbf{Phi4 Few-shot} \\
Beam Search & \cellcolor{bestresult}81.23 & \cellcolor{bestresult}0.12 & \cellcolor{bestresult}0.001 & \cellcolor{bestresult}0.5 & \cellcolor{bestresult}Most reliable \\
Top-$k$ Sampling & 80.28 & 0.20 & 0.002 & 1.1 & Good alternative \\
Top-$p$ Sampling & 80.69 & 0.18 & 0.002 & 0.8 & Balanced performance \\
Temperature & 80.57 & 0.21 & 0.003 & 0.9 & Moderate variance \\
\addlinespace
\textbf{Gemma3 Few-shot} \\
Beam Search & 77.81 & 0.14 & 0.002 & 0.3 & Consistent but lower \\
Top-$k$ Sampling & 77.70 & 0.11 & 0.001 & 0.5 & Very consistent \\
Top-$p$ Sampling & 77.68 & 0.14 & 0.002 & 0.4 & Stable performance \\
Temperature & 77.70 & 0.08 & 0.001 & 0.2 & Most consistent \\
\bottomrule
\end{tabular}
\caption{Decoding strategy robustness and variance analysis. CV = coefficient of variation (Std Dev / Mean), Range = max - min across datasets. Highlighted cells indicate the best combination of performance and stability.}
\label{tab:decoding_robustness}
\end{table}

\subsection{POS Class Performance}
\label{a:pos_class_effect}

\begin{table}[H]
\centering
\begin{tabular}{@{}lccccc@{}}
\toprule
\textbf{POS Class} & \textbf{UDPipe} & \textbf{Phi4 Few-shot} & \textbf{Gemma3 Fine-tuned} & \textbf{Gemma3 CLTF} & \textbf{Improvement} \\
\midrule
ADJ & 54.12 & 58.11 & \cellcolor{bestresult}\textbf{79.75} & 71.94 & +25.63 \\
ADV & 51.19 & 58.79 & \cellcolor{bestresult}\textbf{77.30} & 72.93 & +21.74 \\
PRON & 62.19 & 68.66 & \cellcolor{bestresult}\textbf{84.47} & 81.10 & +22.28 \\
DET & 71.69 & 74.40 & 87.32 & \cellcolor{bestresult}\textbf{89.38} & +17.69 \\
PROPN & 79.90 & 76.26 & \cellcolor{bestresult}\textbf{98.07} & 91.22 & +18.17 \\
\midrule
NOUN & 86.14 & 85.34 & \cellcolor{bestresult}\textbf{91.84} & 88.72 & +5.70 \\
VERB & 91.31 & \cellcolor{bestresult}\textbf{93.55} & 92.29 & 87.91 & +2.24 \\
ADP & \cellcolor{bestresult}\textbf{94.34} & 93.09 & 94.16 & 92.65 & -0.18 \\
CCONJ & 95.02 & 95.86 & \cellcolor{bestresult}\textbf{98.89} & 96.22 & +3.87 \\
\bottomrule
\end{tabular}
\caption{Per-class F1 on \textbf{CAT}: low-performing (top) and high-performing (bottom) classes. Best per row in \colorbox[HTML]{D4EDDA}{\textbf{green}}.}
\label{tab:pos_performance_cat}
\end{table}

\begin{table}[H]
\centering
\begin{tabular}{@{}lccccc@{}}
\toprule
\textbf{POS Class} & \textbf{UDPipe} & \textbf{Phi4 Few-shot} & \textbf{Gemma3 Fine-tuned} & \textbf{Gemma3 CLTF} & \textbf{Improvement} \\
\midrule
NUM & 48.78 & 60.87 & 83.33 & \cellcolor{bestresult}\textbf{88.04} & +39.26 \\
AUX & \cellcolor{bestresult}\textbf{56.45} & 32.97 & 40.00 & 49.30 & -7.15 \\
ADJ & 68.66 & 64.52 & 57.14 & \cellcolor{bestresult}\textbf{70.27} & +1.61 \\
\midrule
ADV & 75.59 & \cellcolor{bestresult}\textbf{77.83} & 64.15 & 74.02 & +2.24 \\
PROPN & 76.19 & 66.67 & 75.00 & \cellcolor{bestresult}\textbf{90.47} & +14.28 \\
VERB & 86.32 & 82.55 & 67.39 & \cellcolor{bestresult}\textbf{86.71} & +0.39 \\
PRON & \cellcolor{bestresult}\textbf{88.50} & 76.66 & 83.72 & 79.90 & -4.78 \\
DET & \cellcolor{bestresult}\textbf{91.79} & 82.72 & 76.47 & 89.25 & -2.54 \\
NOUN & \cellcolor{bestresult}\textbf{92.88} & 90.87 & 89.51 & 88.29 & -2.01 \\
ADP & \cellcolor{bestresult}\textbf{93.14} & 87.70 & 90.62 & 92.16 & -0.98 \\
CCONJ & 93.99 & 89.42 & 91.30 & \cellcolor{bestresult}\textbf{95.65} & +1.66 \\
\bottomrule
\end{tabular}
\caption{Per-class F1 on \textbf{Chauliac}: low-performing (top) and high-performing (bottom) classes. Best per row in \colorbox[HTML]{D4EDDA}{\textbf{green}}.}
\label{tab:pos_performance_chauliac}
\end{table}

\section{Decoding effects}
\label{a:decoding}

\begin{figure*}[h]
    \centering
    \includegraphics[width=1\textwidth]{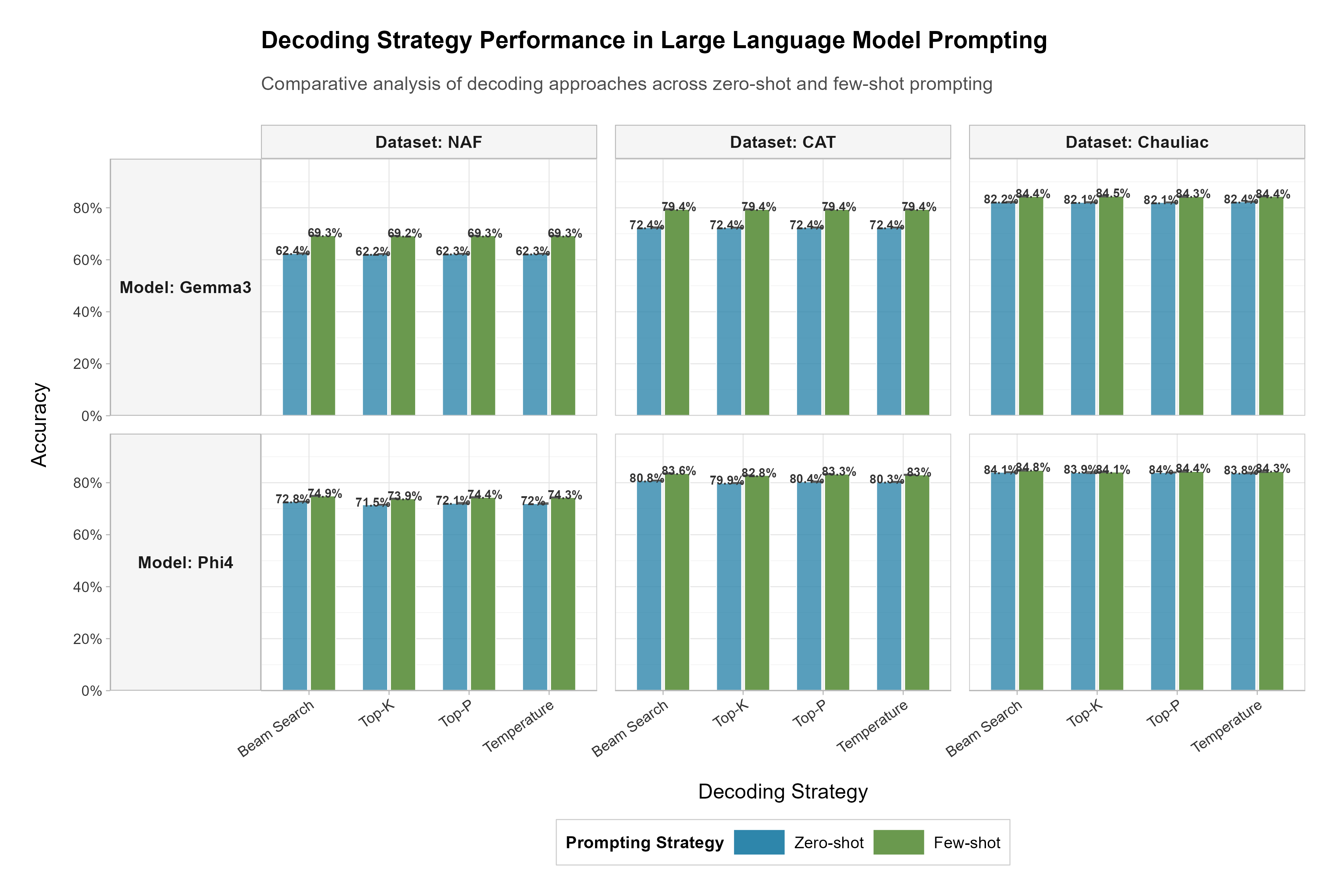}
    \caption{Decoding strategy performance across varying prompts, models and datasets.}
    \label{fig:decoding_performance}
\end{figure*}

\section{Evaluation metrics}
\label{a:metrics}

We assessed our model using several standard metrics, defined as follows.

\paragraph{Accuracy} 
Accuracy quantifies the proportion of tokens for which the predicted POS tag matches the gold label:
\begin{equation}
    \text{Accuracy} = \frac{\#\text{correct predictions}}{\#\text{tokens}}.
\end{equation}

\paragraph{Precision} 
Precision measures the fraction of correct POS tag predictions among all instances predicted as a given tag:
\begin{equation}
    \text{Precision} = \frac{TP}{TP + FP}.
\end{equation}

\paragraph{Recall} 
Recall determines the proportion of actual POS tag instances that were correctly predicted:
\begin{equation}
    \text{Recall} = \frac{TP}{TP + FN}.
\end{equation}

where \(TP\), \(FP\), and \(FN\) denote true positives, false positives, and false negatives, respectively.

\paragraph{F1-score} 
The F1-score, representing the harmonic mean of precision and recall, is computed as:
\begin{equation}
    \text{F1-score} = 2 \times \frac{\text{Precision} \times \text{Recall}}{\text{Precision} + \text{Recall}}.
\end{equation}

\section{Practical Recommendations}
\label{sec:practical_recommendations}

\subsection{Method Selection Framework}

Performance analysis reveals distinct optimal strategies depending on computational resources and target language characteristics, as illustrated in Table \ref{tab:method_selection}.

For resource-scarce languages like Medieval Occitan, trilingual CLTF provides substantial gains (+21.67 percentage points over traditional methods). Medieval Catalan benefits most from dedicated fine-tuning, while Medieval French achieves peak performance through bilingual pairing with Catalan, surpassing both trilingual CLTF and traditional tools. Under resource constraints, UDPipe remains a competitive option for Medieval French.

\subsection{Implementation Guidelines}

\paragraph{Prompting Configuration}
Few-shot prompting consistently outperforms zero-shot across all conditions, with improvements ranging from 2.94 to 10.24 percentage points. Phi4 demonstrates superior prompting capabilities, achieving 81.23\% average accuracy compared to Gemma3's 77.81\%. For decoding, beam search with width 15 provides optimal results across all datasets and models.

\paragraph{Cross-Lingual Transfer Learning}
CLTF shows particular effectiveness for under-resourced varieties. Medieval Occitan achieves the largest improvement (+9.59 percentage points over monolingual fine-tuning with trilingual CLTF), while Medieval French benefits most from targeted bilingual pairing with Catalan (+9.50 percentage points over monolingual fine-tuning). We recommend trilingual CLTF when the target language can benefit from broad cross-lingual exposure, and bilingual CLTF when a closely related language with sufficient training data is available—particularly for small target datasets where trilingual training may dilute the transfer signal.

\paragraph{Strategic Language Pairing}
Our bilingual experiments demonstrate that linguistic similarity should guide language pairing decisions. Catalan, occupying an intermediate position between Gallo-Romance and Occitano-Romance, serves as an effective bridge language for both Occitan and French. In contrast, the French--Occitan pairing yields minimal gains (+0.22 percentage points), suggesting that more distant genealogical relationships or domain mismatches can limit transfer effectiveness. We recommend practitioners prioritize pairings based on genealogical proximity and complementary corpus characteristics rather than defaulting to the largest possible training set.

\paragraph{Performance-Cost Trade-offs}
The progression from prompting (77.75\% average accuracy) to fine-tuning (85.19\%) to CLTF (up to 93.14\% with optimal pairing) represents varying returns relative to computational investment. For production systems processing single languages, the 7.44 percentage point improvement from prompting to fine-tuning may justify computational costs. Bilingual CLTF offers an attractive middle ground: it requires only one additional language's training data yet can deliver substantial gains, particularly for small target datasets. Trilingual CLTF provides the most consistent performance across all languages but may be suboptimal for individual targets where a strong bilingual pairing exists.

\subsection{Quality Assurance Considerations}

Error analysis reveals systematic performance patterns across POS classes. Content words (ADJ, ADV, PRON) show the largest improvements with neural methods, with F1-score gains exceeding 25 percentage points for adjectives and adverbs. Function words (ADP, CCONJ) maintain consistently high performance (>90\% F1) across all methods, suggesting reliable baseline capabilities. For production deployment, we recommend implementing class-specific validation protocols, particularly for content word categories where traditional methods show substantial limitations (ADJ: 54.12\% F1 with UDPipe vs. 79.75\% with fine-tuned models).

\subsection{Resource Allocation Strategy}

Based on performance variance analysis, trilingual CLTF provides the most stable results across cells (sample SD = 1.31 percentage points, CV $\approx 0.015$), whereas traditional methods show high variability (sample SD $\approx 13.0$ percentage points across cells). For multi-language digital humanities projects, trilingual CLTF training followed by language-specific evaluation provides robust performance with predictable resource requirements. However, when the goal is to maximize accuracy for a specific target language, practitioners should evaluate bilingual pairings with genealogically proximate languages before defaulting to trilingual training.

\begin{table*}[htbp]
\centering
\begin{tabular}{@{}lll@{}}
\toprule
\textbf{Dataset} & \textbf{High Resources} &  \textbf{Limited Resources} \\
\midrule
NAF (Medieval Occitan) & Trilingual CLTF (89.68\% acc.) & Few-shot Prompting (75.01\% acc.) \\
CAT (Medieval Catalan) & Fine-tuning (92.52\% acc.) & Few-shot Prompting (83.69\% acc.) \\
Chauliac (Medieval French) & Bilingual CAT+FR (93.14\% acc.) & UDPipe (89.40\% acc.) \\
\bottomrule
\end{tabular}
\caption{Method selection by dataset and computational constraints.}
\label{tab:method_selection}
\end{table*}

\end{document}